\documentclass{article}
\usepackage{ijcai24}

\usepackage{times}
\usepackage{soul}
\usepackage{url}
\usepackage[hidelinks]{hyperref}
\usepackage[utf8]{inputenc}
\usepackage[small]{caption}
\usepackage{graphicx}
\usepackage{amsmath}
\usepackage{amssymb}
\usepackage{amsthm}
\usepackage{booktabs}
\usepackage{algorithm}
\usepackage{algorithmic}
\usepackage[switch]{lineno}
\usepackage{hyperref}
\usepackage{multirow}

\title{Enhancing Boundary Segmentation for Topological Accuracy with Skeleton-based Methods}

\author{
Chuni Liu$^1$\thanks{~indicates equal contribution.}
\and
Boyuan Ma$^1 {}^*$
\thanks{~corresponding authors: \{mbytony, banxj, xuke\}@ustb.edu.cn} \and
Xiaojuan Ban$^1 {}^2 {}^\dagger$ \and
Yujie Xie$^1$\and
Hao Wang$^1$\and
Weihua Xue$^3$\and
Jingchao Ma$^1$\and
Ke Xu$^1{}^\dagger$\\
\affiliations
\small
$^1$Collaborative Innovation Center of Steel Technology, Beijing Advanced Innovation Center for Materials Genome Engineering, School of Intelligence Science and Technology, Shunde Innovation School, Institute for Advanced Materials and Technology, Key Laboratory of Intelligent Bionic Unmanned Systems, University of Science and Technology Beijing, Beijing 100083, China\\
$^2$Institute of Materials Intelligent Technology, Liaoning Academy of Materials, Shenyang 110004, China\\
$^3$School of Materials Science and Technology, Liaoning Technical University, Liaoning 114051, China\\
\emails
\small
chuniliu@xs.ustb.edu.cn
}

\begin{document}

\maketitle

\begin{abstract}
Topological consistency plays a crucial role in the task of boundary segmentation for reticular images, such as cell membrane segmentation in neuron electron microscopic images, grain boundary segmentation in material microscopic images and road segmentation in aerial images. In these fields, topological changes in segmentation results have a serious impact on the downstream tasks, which can even exceed the misalignment of the boundary itself. To enhance the topology accuracy in segmentation results, we propose the Skea-Topo Aware loss, which is a novel loss function that takes into account the shape of each object and topological significance of the pixels. It consists of two components. First, a skeleton-aware weighted loss improves the segmentation accuracy by better modeling the object geometry with skeletons. Second, a boundary rectified term effectively identifies and emphasizes topological critical pixels in the prediction errors using both foreground and background skeletons in the ground truth and predictions. Experiments prove that our method improves topological consistency by up to 7 points in VI compared to 13 state-of-art methods, based on objective and subjective assessments across three different boundary segmentation datasets. The code is available at \href{https://github.com/clovermini/Skea_topo}{https://github.com/clovermini/Skea\_topo}.
\end{abstract}

\section{Introduction}

\begin{figure}[t]
\centering
\includegraphics[width=0.9\columnwidth]{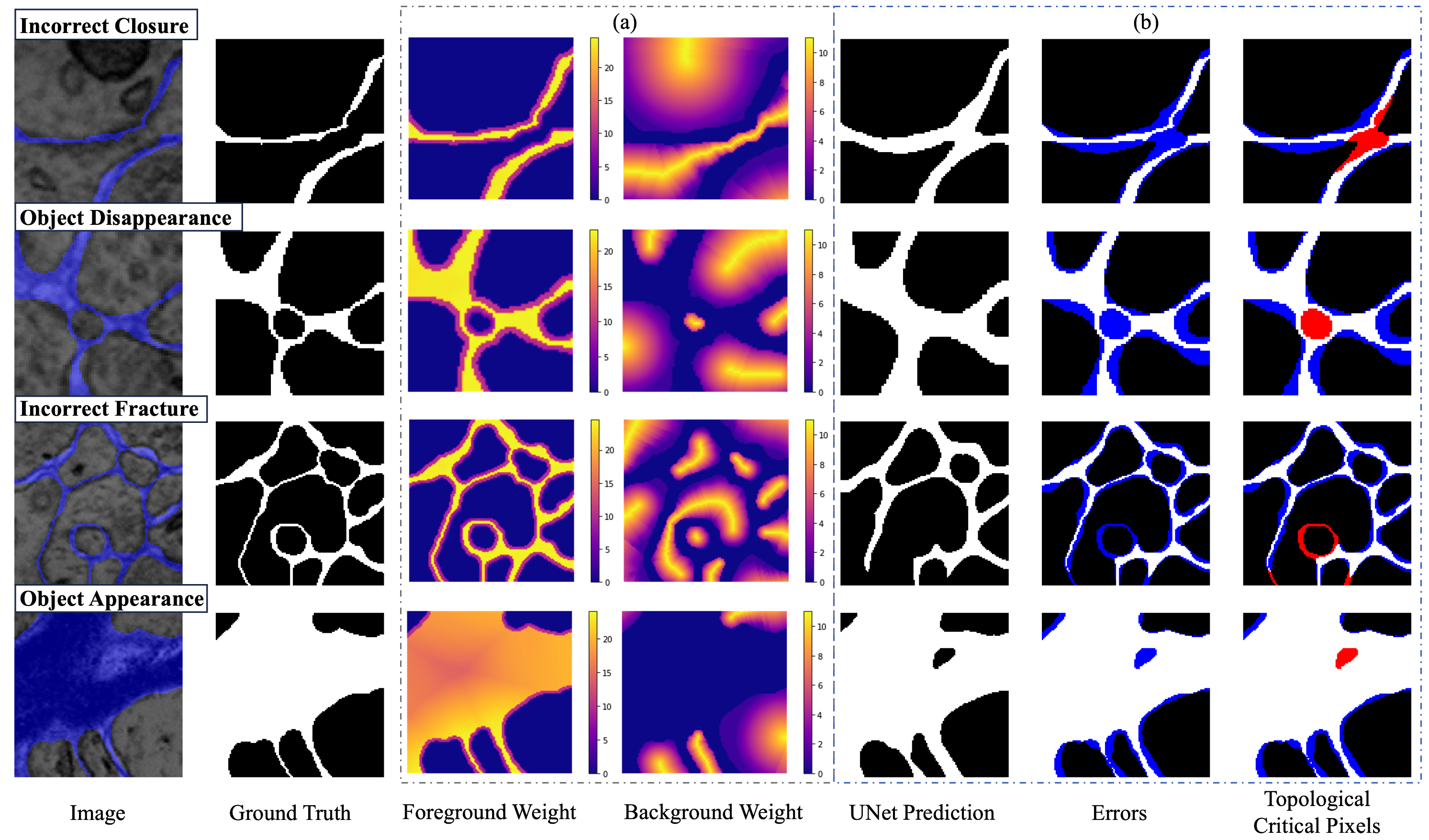} 
\caption{Examples of four types of errors that cause topological changes in boundary segmentation results: Incorrect Closure, Object Disappearance, Incorrect Fracture, and Object Appearance. (a) Foreground and background weighted maps calculated by Skeaw. The background weight accurately models the geometric features of each object. (b) Illustration of the topological critical pixels identified by BoRT. UNet Prediction is the prediction result of the UNet model trained with the cross-entropy loss. The blue areas represent the discrepancies between prediction and ground truth, which signify errors. The red areas highlight the topological critical pixels, which constitute only a small portion of these errors.}
\label{fig1}
\end{figure}

Although deep learning algorithms for boundary segmentation show high accuracy in pixel-level metrics, achieving topologically consistent predictions that align with the ground truth remains a significant challenge. This is particularly evident in the boundary segmentation of reticular images, such as neuron images \cite{conn_res}, material microscopic images \cite{abw} and aerial images \cite{volodymyr:road}. Topologically consistent segmentation demands the preservation of key attributes such as the connectivity, number of holes, and layout of segments. These attributes are crucial for the accuracy of subsequent analyses such as studying the structure and connectivity of neuronal circuits, quantitatively characterizing the microstructure, and navigation planning \cite{anil-et-al:navigation}. 
In boundary segmentation, errors that cause topological changes can be divided into four types, as illustrated in Fig. \ref{fig1}. When viewed in terms of objects such as grains and cells, these errors manifest as the misclassification of critical pixels, which results in unusual appearances, unusual disappearances, and object merging or splitting.

Even a small number of misclassified pixels in the prediction result can lead to these errors with significant topological changes. Thus, standard pixel-based loss functions such as the cross-entropy (CE) loss and Dice loss exhibit inadequate topological correctness. These methods evaluate each pixel equally regarding the classification precision, so the overall loss can reach a low value even when the topological structure is incorrect. Some approaches such as \cite{abw,rrw} assess the misclassification severity based on the location of each pixel and adaptively weight each pixel within the objects (background) according to their distance from the boundary (foreground), as dictated by a distance function. This technique enhances the accuracy and topological consistency of the segmentation. Nevertheless, existing distance-based weighting methods struggle to accurately represent irregular object shapes and may fail to appropriately allocate greater weight to error-prone areas, such as narrow regions in irregular objects. In addition, they are unaware of whether the segmentation results have topological errors. Methods such as \cite{topo,warp} focus on identifying topological critical pixels during training based on network predictions. These are specific pixels in binary images whose value reversal would cause topological changes. These methods place a greater emphasis on the accurate classification of topological critical pixels, which significantly enhances the topological consistency of the segmentation results. However, they currently suffer from noise, a lack of optimal penalty strategies, and time-consuming computations.

In this work, we propose a novel loss function called Skea-Topo Aware to address the aforementioned challenges. Our approach incorporates two key components: a skeleton-aware weighted loss (Skeaw) and a boundary rectified term (BoRT).

Skeaw introduces an innovative weighting approach that applies to both foreground and background pixels, considering the importance of each pixel based on the location and overall category quantity ratio. For the foreground, the weight is calculated based on the overall boundary thickness: a thinner boundary is assigned a larger weight, since it is more fragile and prone to fracture. 
For the background, Skeaw enhances the existing distance-weighting function by employing the object's skeleton to refine the modeling of its shape, which ensures that the weighting results adequately account for the shape features of irregular objects. The proposed skeleton-based approach leverages the benefits of existing distance-weighted methods while overcoming their limitations for irregular objects.

The core motivation of BoRT is to more rapidly identify the topological critical pixels in errors and apply superior penalty methods to steer the model towards generating topologically consistent results. An in-depth analysis of the error regions that cause topological changes has led to the development of a simpler and more efficient method, which serves as an alternative to the existing time-consuming identification techniques. 
This method identifies topological critical pixels by leveraging the trait that they consistently coincide with the foreground or background skeleton in the ground truth and predictions. Conversely, non-topological critical pixels typically do not intersect with these skeletons, except for errors distant from the boundaries, which also necessitate increased penalties. By using the foreground and background skeletons, BoRT can accurately identify connected regions, where topological critical pixels are located, through a process of high computational efficiency. Based on this identification, a novel penalty term, which assigns higher penalties to critical pixels and reduces the penalty strength for non-critical pixels, is devised using likelihood maps to effectively improve the topological consistency of the segmentation result.

In general, our contributions are highlighted as follows:
\begin{itemize}

\item We propose a skeleton-aware weighted loss, which better utilizes the geometric information of objects to establish a weighted map. This approach is more conducive to preserving the geometric and topological properties of irregular objects during the learning process.

\item We propose a novel boundary rectified term to efficiently and effectively identify and emphasize topological critical pixels that affect the topology during training. This term can guide the network to effectively preserve the topology of the boundary segmentation results.

\item We conduct extensive experiments for three different boundary segmentation tasks: road segmentation, cell membrane segmentation, and grain boundary segmentation. Our method performs better objective and subjective assessments than the state-of-the-art (SOTA) methods. In addition, the boundary rectified term can be used as a plug-in for other classical loss functions to further improve the performance.

\end{itemize}

\section{Related work}

In recent years, deep learning-based boundary segmentation methods have undergone rapid development. By introducing the fully convolutional layer, \cite{bcefcn} enables pixel-level image classification and establishes the foundation for a deep learning image segmentation model. Many methods have since optimized the model to improve the performance and accuracy from various angles. The UNet model, which was introduced by \cite{unet}, presents an elegant symmetrical encoding and decoding structure. It incorporates residual connections in the middle layers, which effectively preserve intricate details from the original image. UNet and its variants \cite{unet++,nnunet,dilated_unet,mednext} have gained widespread popularity due to their excellent and robust performance in boundary segmentation tasks. Our work takes UNet as a baseline to assess different loss functions, and our approach is adaptable to various network architectures.

For deep learning models that are optimized using a gradient descent \cite{ce}, a common approach is to adjust the learning direction of the network by designing a target-based loss function. To preserve the topology, various loss functions have been proposed to directly or indirectly increase the penalty when topological errors occur.

\subsubsection{Weighted Losses}
Certain losses prioritize weighting areas that are prone to topological errors, which emphasizes the segmentation correctness on boundaries. For example, \cite{bcefcn} introduced a class balance cross-entropy loss (BCE) that applied heavier weights to boundary pixels based on the number of categories. The Boundary loss \cite{boundary} calculates the loss based on boundary errors only. GraphCuts loss \cite{graph_cuts} add boundary penalty term based on the graph cuts theory. Dice loss \cite{dice}, Tversky loss \cite{tversky}, Lovasz loss \cite{lovasz}, and margin loss \cite{margin} collectively treat pixels of the same class and emphasize the overall boundary segmentation accuracy. These methods overlook the varying topological importance of pixels located at different positions. UNet \cite{unet} introduces a weighted cross-entropy loss (WCE), which considers the distance between boundary pixels and nearby objects. However, for evenly spaced objects, it is equal to the BCE. Some losses weight the pixels in each object based on a distance transform function, such as the error-penalizing distance weighted loss (EPDW) \cite{epdw}, adaptive boundary weighted loss (ABW) \cite{abw}, and region-wise loss (RRW) \cite{rrw}. However, their weighting methods are not suitable for irregular objects, which have narrow or curved regions that are difficult to classify. These regions have lower weights because their pixels are close to the boundary. Our method solves this problem by normalizing the weights using the skeleton.

\subsubsection{Topology-Aware Losses}
Many methods more directly improve the model's perception of the topological structure of images to enhance the topological consistency of segmentation results. Topoaware loss \cite{topoaware} and RMI loss\cite{rmi} model the output structure and enforce its consistency, but this structure may be unrelated to the topology. clDice loss \cite{cldice} and DMT loss \cite{dmt} define the skeleton of the boundary as the topological critical structures and enforce these locations to be correctly segmented. However, these pixels may not be the ones that actually cause topology errors. \cite{malis,mala} adopt the idea of the rand index, find errors in the affinity graph that caused split and merge errors by calculating the maximin path among pixels and imposed strong penalties. Based on this method, \cite{malis_window} focuses on the connectivity of background regions in the predicted distance map and penalizes the unwanted interruptions and false connections. However, the computational complexity of comparing pixel pairs is very high, and focusing on a single pixel with the smallest value can lead to unstable training. Many methods use persistent homology-based methods to describe and compare the topological structures of images. \cite{topo,ph-topo} compare the persistence diagrams to identify critical pixels in the likelihood map, but they can be easily affected by noise. \cite{betti_match} uses induced matchings among persistence barcodes to precisely formalize the spatial correspondences between topological structures of two images. However, the computational complexity of persistent homology methods is cubic in relation to image size, making them time-consuming for large datasets. \cite{warping} proposes a warping error method, which can accurately determine whether a pixel can cause a topological change based on the digital topological knowledge. Warp loss\cite{warp} accelerates it through the proposed efficient Distance-Ordered Homotopy Warping method, which identifies critical pixels through homotopic warping pixels on the discrepancies between predicted results and ground truth in order of the distance. However, it requires a pixel-wise operation, and its penalty method is not optimal. In our work, we propose a much more efficient approach to accurately identify topological critical pixels and design a better penalty term.

\begin{figure*}[t]
\centering
\includegraphics[width=0.8\textwidth]{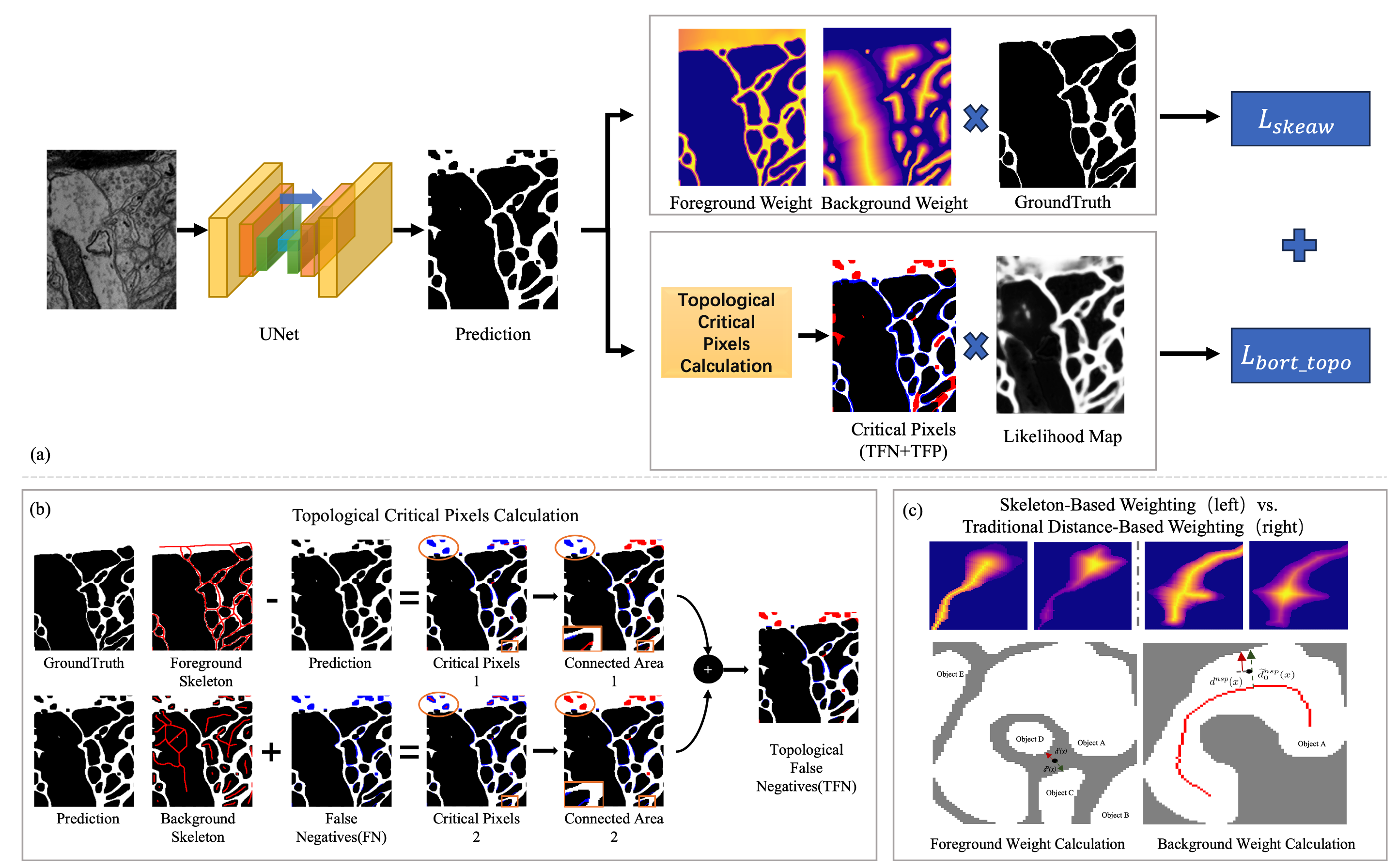} 
\caption{
(a) Illustration of the proposed Skea-Topo Aware Loss. (b) Flowchart of the process to calculate topological false negatives. Our method utilizes both foreground and background skeletons from the ground truth and predicted results, which complement each other to precisely identify the critical pixels. (c) Results of Skeleton-Based Weighting versus Traditional Distance-Based Weighting and calculations for $d^1(x)$, $d^2(x)$, $d^{nsp}(x)$ and $\widetilde{d}_0^{nsp}(x)$. Compared to traditional weighted methods, the skeleton-based approach assigns higher weights to the narrow regions in irregular objects.}
\label{fig2}
\end{figure*}

\section{Method}
The proposed Skea-Topo Aware loss consists of a Skeleton Aware Weighted loss and a Boundary Rectified Term (as shown in Fig. \ref{fig2} (a)). The former precomputes the weight map before training. The latter identifies topological critical pixels during training. These two components complement each other and collaborate in the overall process. Therefore, the total loss function is defined as follow:
\begin{equation}
\label{formula1}
L_{total} = L_{skeaw}+\lambda L_{bort\_topo}
\end{equation}
where $L_{skeaw}$ is the Skeleton Aware Weighted loss, $L_{bort\_topo}$ is the Boundary Rectified Term, and the weight $\lambda$ is utilized to balance the two terms.

\subsection{Skeleton Aware Weighted loss}
By using a combination of a pixel-wise weight map and the cross-entropy loss, the attention of the network towards specific pixels can be regulated. For the boundary segmentation task in this work, each pixel is classified as an object (background, class 0) or a boundary (foreground, class 1). The Skeleton Aware Weighted loss is formulated as follows:

\begin{equation}
\begin{aligned}
\label{formula2}
& L_{skeaw} =\\ -\sum_{x \in \Omega}
& \begin{cases} w_1^{s}(x)log(p_1(x)), & \mbox{if }w_0^{s}(x) < w_1^{s}(x) \cdot m_{d}(x) \\ w_0^{s}(x) log(p_0(x)), & \mbox{if }w_0^{s}(x) \geq w_1^{s}(x) \cdot m_{d}(x)  \end{cases}
\end{aligned}
\end{equation}
where $p_0(x)$ and $p_1(x)$ are the probabilities obtained after applying the softmax activation function. $w_1^{s}(x)$ and $w_0^{s}(x)$ are the weighted values assigned to each pixel $x$ $\in$ $\Omega$($\Omega \subset \mathbb{R}^{2,3}$) in the foreground and background, respectively. $m_d(x)$ is the dilation mask of the ground truth, the actual foreground and background are determined by comparing the weight of the background with that of the dilated foreground. This approach increases the tolerance for slight offsets between predicted and ground truth boundaries, which can improve the segmentation accuracy. The degree of dilation is controlled by the hyperparameter $d_{iter}$. 

The design of our loss is based on two principles. First, for the boundary, higher weights should be assigned to pixels in narrower boundaries, since they have a higher risk of fracture. Second, for the object, higher weights are assigned to misclassified pixels far away from the boundary, since those near it can be difficult to distinguish, often due to boundary ambiguity and manual annotation misalignment. Existing weighting methods use the distance transform function to assign weights to individual objects. For the stability of training, these methods normalize the distance value using the maximum distance, so relatively low weights are assigned to narrow areas of irregular objects, which are prone to boundary closure errors. To overcome this limitation, our method introduces the object skeleton for distance calculation and weight normalization. The skeleton of an object is extracted through the morphological skeletonization function provided by scikit-image\cite{scikit-image}. Through the object skeleton, a uniform weight distribution can be achieved in narrow areas and other areas, as shown in Fig. \ref{fig2} (c). The detailed definitions of $w_1^{s}(x)$ and $w_0^{s}(x)$ are as follows:

\begin{equation}
\label{formula3}
w_1^{s}(x) = w_1^{bce}(x) + w_0 \times (\frac{2\widetilde{d}_1^{max}-d^1(x)-d^2(x)}{2\widetilde{d}_1^{max}}+1)
\end{equation}

\begin{equation}
\label{formula4}
w_0^{s}(x) = w_0^{bce}(x) + w_0 \times (\frac{d^{nsp}(x)}{\widetilde{d}_0^{nsp}(x)})
\end{equation}
where $d^1(x)$ and $d^2(x)$ are the euclidean distance from boundary pixel $x$ to the closest and next closest objects, respectively. $\widetilde{d}_1^{max}$ is the maximum value among all $d^1(x)$. $d^{nsp}(x)$ is the distance from pixel $x$ within an object to the nearest boundary pixel, and $\widetilde{d}_0^{nsp}(x)$ is the distance from the above nearest boundary pixel to the closest skeleton point of pixel $x$, as shown in Fig. \ref{fig2} (c). The weights $w_1^{bce}(x)$ and $w_0^{bce}(x)$ serve as class-balance weights and are equivalent to the inverse proportion of each category. $w_0$ is a hyperparameter to assign the weight of the weight map, which is set to 10 by default in our experiments. 

\subsection{Boundary Rectified Term}
Before introducing the algorithm, several concepts are defined for better understanding. Positive ($p$) refers to the pixels predicted as the boundary; negative ($n$) refers to the pixels predicted as objects; true ($t$) signifies matching predicted and real classes; false ($f$) signifies mismatched predicted and real classes. True positives ($tp$) and true negatives ($tn$) represent correct predictions; false positives ($fp$) and false negatives ($fn$) indicate incorrect predictions. The errors in this work are the combination of $fp$ and $fn$. Topological critical pixels are equal to the sum of topological false positives ($tfp$) and topological false negatives ($tfn$) and abbreviated as true errors ($tf$), which mean errors that lead to topological changes. Then, $ffp$ and $ffn$ are non-topological critical errors, which do not affect the topology, and abbreviated as false errors ($ff$).

Fig. \ref{fig2} (b) shows the step-by-step legend of the calculation process of $tfn$. Considering the four situations that lead to topological changes, the topological critical pixels that correspond to boundary incorrect closure and object disappearance are only found in $fp$, whereas those linked to boundary incorrect fracture and object appearance only occur in $fn$. Taking these two situations in $fn$ as examples for analysis, when a new object appears in the prediction, its skeleton must intersect with $fn$, so the corresponding connected region of critical pixels can be found by calculating the intersection of the background skeleton of prediction and $fn$. The background skeleton is derived from the sum of the skeletons of each object. The background skeleton can also be used to identify most incorrect fracture regions. Typically, a fractured area becomes a narrow passageway that connects two merged objects, and the background skeleton invariably crosses this bridge area. However, in certain extreme cases such as the one highlighted in the yellow box in Fig. \ref{fig2} (b), the background skeleton fails to detect topological critical pixels. In these instances, the non-overlapping area between the foreground skeleton of the ground truth and the prediction accurately pinpoints the fractured region. This foreground skeleton method does not effectively identify new object regions, as demonstrated in the yellow circle in Fig. \ref{fig2} (b). When these two approaches are used together, they complement each other and collectively cover all potential scenarios. This synergy enables the precise identification of all topological critical pixels. For disappearing objects and incorrect closed boundaries, the same logic applies when the labels are treated as predictions and vice versa.

Skeletons are used because they can effectively exclude most of the $ff$ near the boundary but not areas that significantly alter the shape or are located far from the boundary. Such errors should also be assigned a heavier penalty to maintain the accuracy of the results. Instead of individually examining each pixel to determine whether flipping that pixel will change the topology as Warp\cite{warp}, our approach identifies all critical pixels using a few image-level operation steps through the skeletons. The most time-intensive step of this process is computing the foreground and background skeletons, each with a time complexity of O(n), where n represents the number of image pixels. Leveraging parallelization and hardware acceleration technologies in Python, our method ensures fast execution speeds in practical applications.

Our method directly multiplies the probability map with the identified critical pixels to define the penalty term. The key goal is to enable $ffn$ and $ffp$ to be classified into opposing categories while ensuring the accurate classification of $tp$, $tn$, $tfn$, and $tfp$. Essentially, this step adjusts the actual training target based on the detected $tf$ and $ff$ during training. Therefore, this method is named the boundary rectified term. The loss function is calculated using the rectified likelihood map $p_{l(x)}^{fix}(x)$ (where $l \rightarrow{\{0,1\}}$), and l(x) is the class of pixels), which is derived from the following formula:

\begin{equation}
\label{formula5}
L_{bort\_topo} = \sum_{x \in \Omega}p_{l(x)}^{fix}(x)
\end{equation}

\begin{equation}
\label{formula6}
p_0^{fix}(x) = tn \times p_1(x) + tfp \times p_1(x) + ffp \times p_0(x)
\end{equation}
\begin{equation}
\label{formula7}
p_1^{fix}(x) = tp \times p_0(x) + tfn \times p_0(x) + ffn \times p_1(x)
\end{equation}
where $p_{l(x)}$ is the likelihood map. While not explicitly stated in the formula, $tfp$ and $tfn$ can be properly weighted according to the data situation to obtain better results. Our formula is similar to Poly1\cite{polyloss}. Poly1 adjusts the importance of the first polynomial coefficients of cross-entropy loss, which results in higher segmentation accuracy. A comparison experiment between Poly1 and BoRT will be conducted to further validate the effectiveness of identifying topological critical pixels.

\section{Experiments}
Extensive experiments were conducted to validate the effectiveness of our proposed method in preserving the topology. 
All experiments were performed using an NVIDIA GeForce RTX 3090 GPU (24GB Memory) and an Intel(R) Xeon(R) Silver 4210R CPU @ 2.40 GHz.

\begin{table*}[htb]
\tiny
\setlength{\tabcolsep}{0.5mm}
\centering
\begin{tabular}{lrrrrrrrrrrrrrrr}
\toprule
\multirow{2}{*}{Method} & \multicolumn{5}{c|}{SNEMI3D} & \multicolumn{5}{c|}{IRON} & \multicolumn{5}{c}{MASS. ROAD} \\
 & VI$\downarrow$ & mAP$\uparrow$ & ARI$\uparrow$ & Dice$\uparrow$ & \multicolumn{1}{c|}{Betti$\downarrow$} & VI$\downarrow$  & mAP$\uparrow$ & ARI$\uparrow$ & Dice$\uparrow$ & \multicolumn{1}{c|}{Betti$\downarrow$} & VI$\downarrow$ & mAP$\uparrow$ & ARI$\uparrow$ & Dice$\uparrow$ & Betti$\downarrow$ \\ \midrule
CE & 82.84$_{\pm 9.60}$ & 34.55$_{\pm 0.95}$ & 53.43$_{\pm 0.98}$& 73.31$_{\pm 0.21}$ & 27.09$_{\pm 4.13}$ & 43.37$_{\pm 2.58}$ & 53.91$_{\pm 1.46}$ & 69.67$_{\pm 0.81}$ & 88.45$_{\pm 0.24}$ & 34.49$_{\pm 1.61}$ & 122.43 & 22.22 & 39.97 & 68.97 & 33.82 \\
BCE$_{(CVPR,2015)}$ & 46.70$_{\pm 6.16}$& 41.79$_{\pm 2.08}$& 60.70$_{\pm 3.13}$& 74.69$_{\pm 0.72}$ & 22.31$_{\pm 6.48}$ & \textbf{19.74}$_{\pm 1.79}$ & 56.89$_{\pm 2.43}$ & 69.31$_{\pm 1.33}$ & 87.38$_{\pm 0.49}$ & 37.48$_{\pm 8.57}$ & \underline{76.67} & 25.52 & \textbf{68.32} & \underline{75.87} & 19.04 \\
WCE$_{(MICCAI,2015)}$ & 42.30$_{\pm 4.28}$ & 43.24$_{\pm 3.28}$ & 62.81$_{\pm 1.13}$& 75.32$_{\pm 0.94}$ & 20.20$_{\pm 2.69}$ & -- & -- & -- & -- & -- & -- & -- & -- & -- & -- \\
Dice$_{(3DV,2016)}$ & 57.93$_{\pm 5.72}$ & 43.07$_{\pm 0.83}$ & 59.42$_{\pm 1.40}$& 75.73$_{\pm 0.41}$ & 21.04$_{\pm 1.12}$ & 39.04$_{\pm 4.41}$ & 55.03$_{\pm 1.29}$ & 69.95$_{\pm 0.98}$ & 88.42$_{\pm 0.24}$ & 26.97$_{\pm 6.02}$ & 92.76 & 25.42 & 47.57 & 71.02 & 31.86 \\
Lovasz$_{(CVPR,2018)}$ & 74/79$_{\pm 8.60}$& 39.73$_{\pm 1.33}$& 55.84$_{\pm 2.22}$& 75.26$_{\pm 0.71}$ & 25.91$_{\pm 3.91}$ & 44.94$_{\pm 2.95}$ & 45.61$_{\pm 0.25}$ & 63.51$_{\pm 0.72}$ & 86.95$_{\pm 0.16}$ & 26.13$_{\pm 2.53}$ & 122.77 & 22.33 & 42.00 & 70.05 & 37.67 \\
RMI$_{(NIPS,2015)}$ & 46.77$_{\pm 3.15}$ & 44.53$_{\pm 1.47}$ & 62.09$_{\pm 1.58}$& \textbf{76.59}$_{\pm 0.41}$ & 15.76$_{\pm 1.02}$ & 29.96$_{\pm 1.09}$ & 56.92$_{\pm 1.35}$ & 70.25$_{\pm 1.03}$ & 88.21$_{\pm 0.46}$ & 18.35$_{\pm 4.01}$ & 99.13 & 28.03 & 48.47 & 72.55 & 33.14 \\
GraphCuts$_{(ICCV,2021)}$ & 58.04$_{\pm 3.58}$ & 37.84$_{\pm 2.40}$ & 56.05$_{\pm 1.67}$& 73.99$_{\pm 0.89}$ & 31.83$_{\pm 17.6}$ & 38.36$_{\pm 3.79}$ & 55.73$_{\pm 0.96}$ & 70.45$_{\pm 0.67}$ & \textbf{88.71}$_{\pm 0.14}$ & 26.27$_{\pm 5.67}$ & 91.41 & \underline{32.03} & 63.67 & 74.80 & 22.84 \\
clDice$_{(CVPR,2021)}$ & 45.82$_{\pm 2.71}$ & 44.55$_{\pm 1.86}$ & 63.11$_{\pm 0.83}$& 74.91$_{\pm 0.80}$ & \underline{13.76}$_{\pm 1.74}$ & 21.54$_{\pm 1.33}$ & \underline{59.92}$_{\pm 0.79}$ & \underline{72.00}$_{\pm 0.36}$ & 88.55$_{\pm 0.18}$ & \underline{17.72}$_{\pm 0.94}$ & 92.52 & 26.50 & 43.47 & 71.21 & 24.00 \\
Margin$_{(IJCV,2022)}$ & 74.29$_{\pm 9.85}$ & 41.17$_{\pm 1.70}$ & 56.86$_{\pm 2.48}$& 76.25$_{\pm 0.26}$ & 44.29$_{\pm 6.80}$ & 51.93$_{\pm 4.39}$ & 51.12$_{\pm 0.81}$ & 67.86$_{\pm 1.08}$ & 87.96$_{\pm 0.33}$ & 44.73$_{\pm 6.80}$ & 170.61 & 15.40 & 23.28 & 63.91 & 42.96 \\
Warp$_{(NIPS,2022)}$ & 49.92$_{\pm 8.57}$ & 41.50$_{\pm 2.81}$ & 60.38$_{\pm 2.46}$& 75.33$_{\pm 1.18}$ & 25.30$_{\pm 10.6}$ & 29.02$_{\pm 1.96}$ & 58.03$_{\pm 1.41}$ & 70.63$_{\pm 0.83}$ & 88.24$_{\pm 0.31}$ & 18.03$_{\pm 4.31}$ & 81.54 & 28.56 & 56.02 & 73.47 & 24.35\\
ABW$_{(JCS,2022)}$ & \underline{42.28}$_{\pm 3.47}$ & 43.81$_{\pm 2.08}$ & \underline{64.05}$_{\pm 1.78}$& 75.33$_{\pm 0.76}$ & 18.78$_{\pm 5.20}$ & 20.30$_{\pm 0.62}$ & 55.64$_{\pm 0.82}$ & 67.33$_{\pm 0.07}$ & 86.06$_{\pm 0.02}$ & 34.69$_{\pm 5.13}$ & 92.77 & 23.77 & 58.31 & 73.29 & \underline{16.31} \\
Poly$_{(ICLR,2022)}$ & 69.21$_{\pm 3.75}$& 39.19$_{\pm 0.90}$& 56.94$_{\pm 1.23}$& 74.80$_{\pm 0.24}$ & 28.21$_{\pm 3.81}$ & 41.19$_{\pm 3.57}$& 54.75$_{\pm 1.03}$& 70.07$_{\pm 0.96}$ & 88.58$_{\pm 0.24}$ & 31.65$_{\pm 5.00}$ & 118.81 & 21.91 & 44.50 & 70.22 & 35.25 \\
RRW$_{(PR,2023)}$ & 54.45$_{\pm 2.01}$ & \underline{44.74}$_{\pm 0.87}$ & 59.64$_{\pm 1.19}$& 76.08$_{\pm 0.41}$ & 18.61$_{\pm 1.99}$ & 34.58$_{\pm 1.27}$ & 56.14$_{\pm 0.25}$ & 70.41$_{\pm 0.54}$ & 88.39$_{\pm 0.15}$ & 22.23$_{\pm 3.83}$ & 87.90 & 28.78 & 52.85 & 72.81 & 26.61 \\ 
\midrule
Skea-Topo &\begin{tabular}[c]{@{}c@{}}\textbf{36.59$_{\pm1.94}$}\end{tabular} 
&\begin{tabular}[c]{@{}c@{}}\textbf{47.49}$_{\pm 0.88}$\end{tabular} 
&\begin{tabular}[c]{@{}c@{}}\textbf{65.49}$_{\pm 1.21}$\end{tabular} 
&\begin{tabular}[c]{@{}c@{}}\underline{76.43}$_{\pm 0.26}$\end{tabular} 
&\begin{tabular}[c]{@{}c@{}}\textbf{12.72}$_{\pm 2.68}$\end{tabular} 

&\begin{tabular}[c]{@{}c@{}}\underline{19.88}$_{\pm 0.55}$\end{tabular} 
&\begin{tabular}[c]{@{}c@{}}\textbf{60.30}$_{\pm 0.61}$\end{tabular} 
&\begin{tabular}[c]{@{}c@{}}\textbf{72.30}$_{\pm 0.55}$\end{tabular} 
&\begin{tabular}[c]{@{}c@{}}\underline{88.64}$_{\pm 0.26}$\end{tabular} 
&\begin{tabular}[c]{@{}c@{}}\textbf{17.31}$_{\pm 2.85}$\end{tabular} 

& \begin{tabular}[c]{@{}c@{}}\textbf{69.02}\end{tabular} 
& \begin{tabular}[c]{@{}c@{}}\textbf{33.19}\end{tabular} 
& \begin{tabular}[c]{@{}c@{}}\underline{68.11}\end{tabular}
& \begin{tabular}[c]{@{}c@{}}\textbf{76.61}\end{tabular} 
& \begin{tabular}[c]{@{}c@{}}\textbf{15.92}\end{tabular}\\ \bottomrule
\end{tabular}
\caption{Quantitative results of different losses on the SNEMI3D dataset (Neurites EM Images), Pure Iron Grain dataset (Material Microscopic Images, IRON) and Massachusetts Road dataset (Aerial Road Images, MASS. ROAD). The \textbf{bold} numbers denote the best performance in each metric. The \underline{underlined} numbers denote the second-best performance. All reported metrics are multiplied by 100 except for Betti.}
\label{table1}
\end{table*}

\subsection{Experiment Setting}
\subsubsection{Datasets}

Our method was evaluated using three publicly available image segmentation datasets from different domains. The first dataset, SNEMI3D \cite{snemi3d} in biology, consists of electron microscopy (EM) images to segment mouse cortex neurites in 3D. It has a dataset size of $100\times1024\times1024$, and its 2D slices were used here for the 2D boundary segmentation experiments. The second dataset, Pure Iron Grain(IRON)\cite{abw}, focuses on pure iron grain segmentation in material science. It comprises 296 microscopic slices with an image size of $1024\times1024$.  The third dataset, Massachusetts Roads Dataset (MASS. ROAD)\cite{volodymyr:road} from remote sensing, contains 1171 aerial images of Massachusetts, each of which is 1500×1500 pixels. To assess the model robustness, a three-fold cross-validation approach was used for the SNEMI3D and IRON datasets. For the MASS. ROAD Dataset, validation and testing were conducted using its official sets. 

\subsubsection{Baselines and Implementation Details}
In our experiments, we compared our method with 13 SOTA loss functions (Table \ref{table1}). All methods were trained based on the standard PyTorch implementation of the UNet model. To ensure fairness, a consistent configuration was maintained. In the preprocessing stage, all images were normalized based on the mean and standard deviation(std) calculated on each dataset. The data augmentation techniques of random crop, rotation and flip were applied. The input size was set to $512\times512$. During training, the Adam optimizer with default parameters was used, and the StepLR scheduler with a step size of 10 and a decay rate of 0.8 was used to adjust the learning rate. The initial learning rate was set to 1e-4. Each model was trained for 50 epochs with a batch size of 10, and we obtained the best parameter based on early stopping. These parameter values were selected through a grid search. All SOTA loss functions were implemented using its official PyTorch implementation. Poly1 \cite{polyloss} based on the cross-entropy loss was used. Hyperparameter $\alpha$ for clDice \cite{cldice} was set to 0.1 through a grid search. For datasets with evenly spaced objects such as IRON and MASS. ROAD, WCE \cite{unet} is equivalent to BCE \cite{bcefcn}, since all foreground pixels have identical weights.

\subsubsection{Evaluation Metrics}
Five metrics were used in the experiment to evaluate the performance of each method. These metrics include a pixel-wise metric (Dice) and topology-related metrics: Variation of Information (VI) \cite{voi}, Adjusted Rand Index (ARI) \cite{ari}, mean average precision (mAP) \cite{map}, and Betti Error (Betti). VI is given prominence in our evaluation because it is sensitive to the presence of topological changes and the extent of their impact. To ensure fair evaluation, post-processing was applied to the segmentation results and ground truth. It includes three steps: boundary dilation with a square filter of size 3 to remove small holes, skeletonization, and a second round of dilation using the same filter.

\subsection{Qualitative and Quantitative Results}
\begin{figure}[ht]
\centering
\includegraphics[width=0.9\columnwidth]{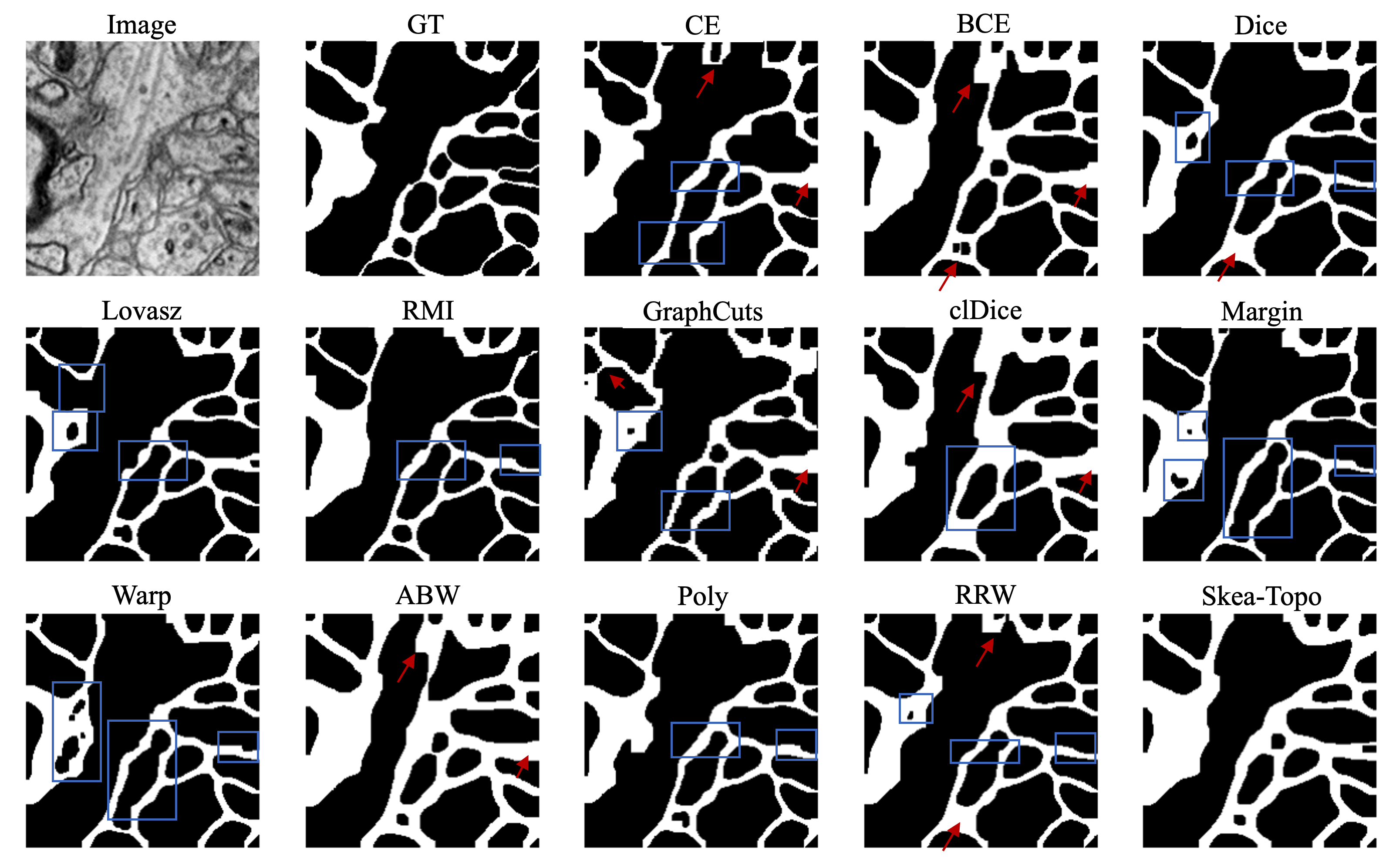}
\caption{Qualitative results of different losses on the SNEMI3D dataset. The red arrows and blue rectangles indicate topological false positive and false negative errors, respectively.}
\label{fig3}
\end{figure}

\begin{figure}[ht]
\centering
\includegraphics[width=0.9\columnwidth]{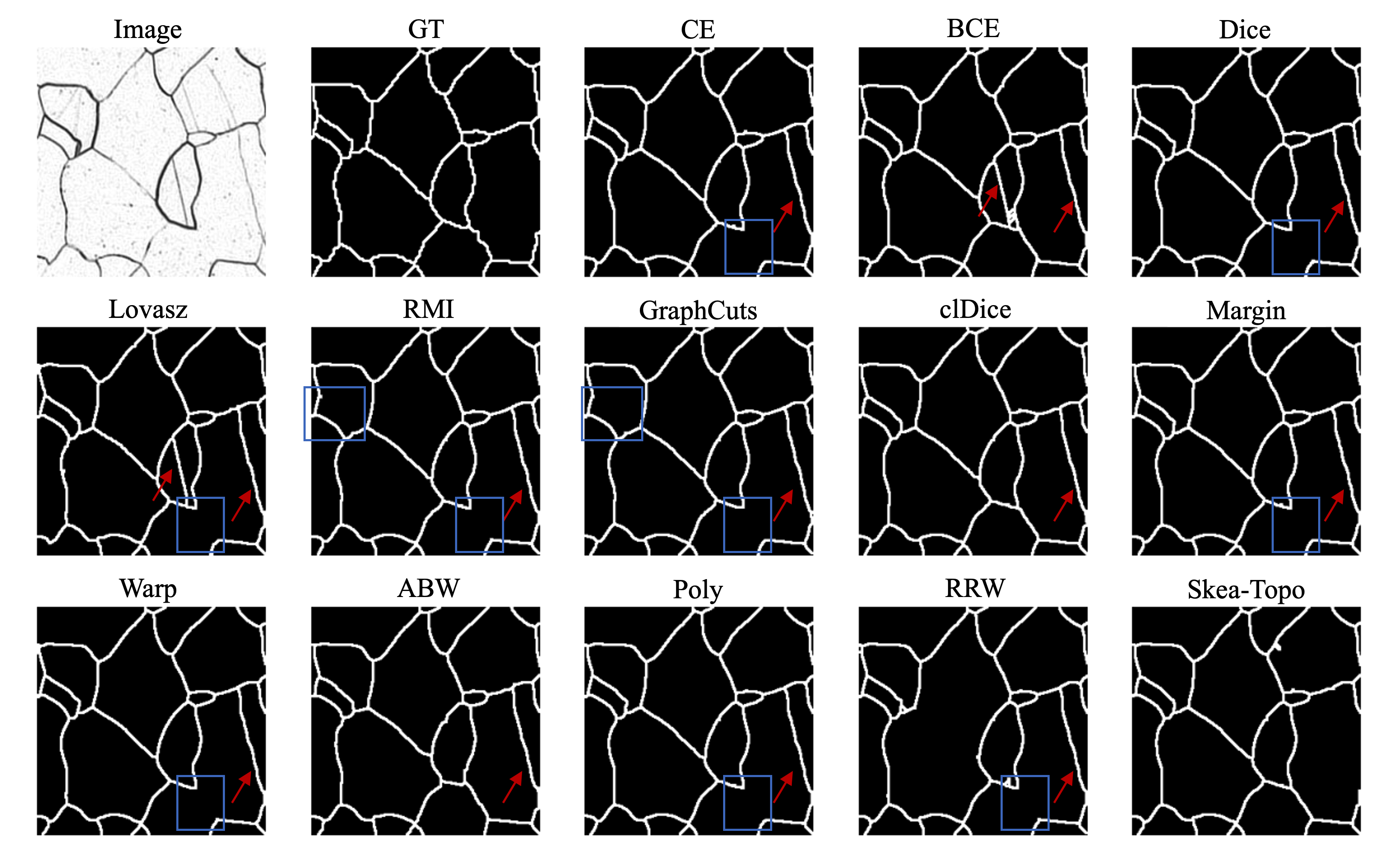}
\caption{Qualitative results of different losses on the IRON dataset. }
\label{fig4}
\end{figure}

\begin{figure}[ht]
\centering
\includegraphics[width=0.9\columnwidth]{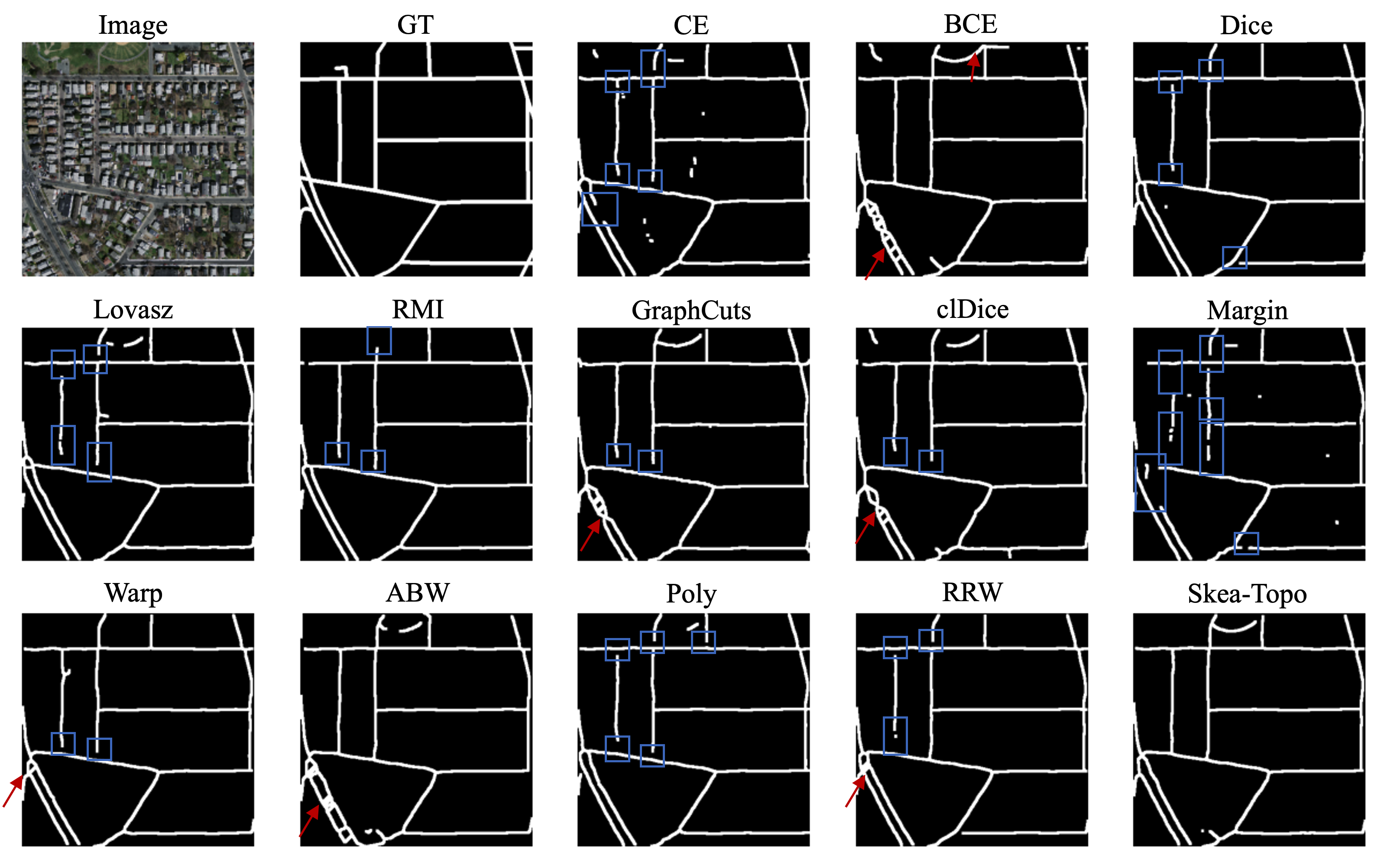}
\caption{Qualitative results of different losses on the MASS. ROAD dataset. }
\label{fig5}
\end{figure}

Table \ref{table1} displays the quantitative results of our loss function compared to 13 other SOTA losses. Our method delivers top-tier performance across most metrics: it ranks first in most cases and a close second in some cases. Other methods exhibit their own strengths and weaknesses on different metrics, whereas our approach excels across all evaluated metrics, which showcases its well-rounded effectiveness. Compared to the existing methods, our approach achieves a maximum improvement of 7 points in VI. Significantly, no other method has simultaneously shown superior performance across all three datasets as ours does, which underscores the exceptional effectiveness and robustness of our approach. The subscript data with $\pm$ represent the std of the results of three-fold experiments. The relatively low std of our method highlights its stability. All reported metrics of our method are based on the best overall performance in the ablation studies. Figs. \ref{fig3}, \ref{fig4}, and \ref{fig5} qualitatively compare the boundary segmentation results between our method and other methods across three datasets. These figures demonstrate that our method is more effective than other methods in eliminating topological errors when compared to the ground truth (GT). 

\subsection{Ablation Studies}
In this section, only the ablation experiment results on the SNEMI3D dataset are reported, and the loss weight $\lambda$ was set to 1 by default, which already significantly improves performance.

\subsubsection{Choice of hyperparameters}
There are two critical hyperparameters involved in the proposed loss function. The first one is the dilation iteration ($d_{iter}$) in Skeaw, which is designed to broaden the coverage of the foreground. A value of 1 in this context indicates a dilation operation using a square filter with a size of 3. The second hyperparameter determines the epoch at which we integrate BoRT into the total loss ($step\_num$). Intuitively, introducing BoRT when the network is in a more stable state instead of the early noisy stages of training should result in better improvements. It should be noted that the hyperparameter ablation experiments for BoRT were conducted based on the optimal configuration of Skeaw.

\begin{figure}[ht]
\centering
\includegraphics[width=0.95\columnwidth]{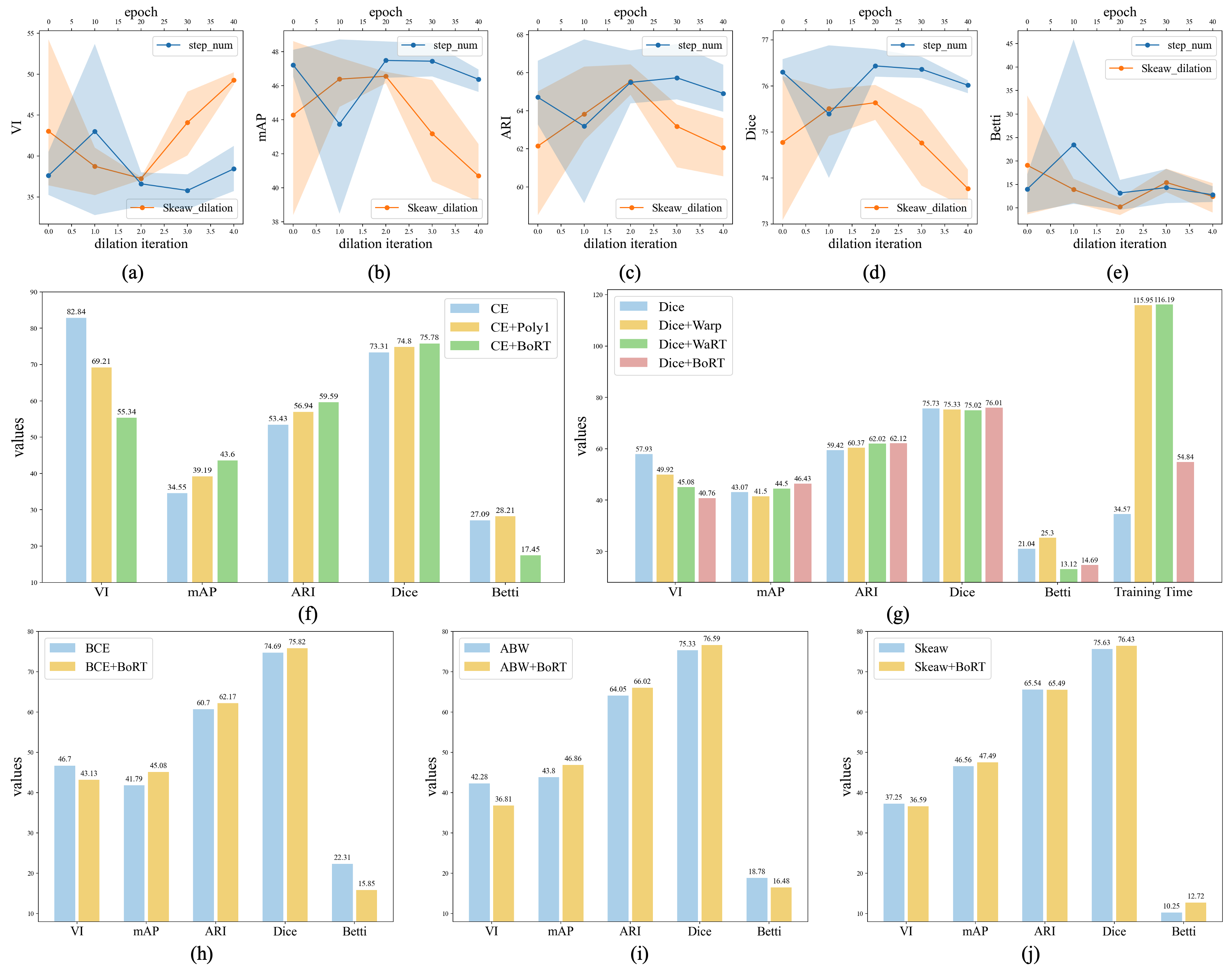}
\caption{(a)-(e) Ablation study for the choice of hyperparameters. (f)-(j) Improvements of five losses after introducing BoRT. (f) Additional comparison with the Poly loss; (g) Additional performance and efficiency comparison with Warp and WaRT.}
\label{fig6}
\end{figure}

Based on the results in Fig. \ref{fig6}, the recommended values for these hyperparameters are 2 and 20. These values yield the best performance and a relatively low standard deviation, which indicates stable training. The findings suggest that dilating the ground truth can improve the segmentation accuracy. The trends in Fig. \ref{fig6} also reveal that the effectiveness of BoRT is closely tied to the timing of its inclusion, notably, setting the parameter to 10 leads to a noticeable decrease in performance. In subsequent experiments, $step\_num = 20$ is used as the default setting.

\subsubsection{Impact of the loss terms}
Ablation experiments on each component of our proposed loss were performed to verify their necessity. 
Specifically, the components of the penalty term in BoRT were evaluated, w/o $ff$ refers to removing the $ffp$ and $ffn$ terms in $L_{bort\_topo}$, and w/o $ff$ indicates removing the $tn$ and $tp$ terms. The results in Table \ref{table2} demonstrate the necessity of all penalty components. Furthermore, similar ablation studies were conducted on the other two datasets. The IRON and SNEMI3D datasets have generally consistent conclusions, but the standard form of the penalty term poorly performs on the MASS. ROAD dataset, possibly due to the extreme data imbalance in the MASS. ROAD dataset. Consequently, we modified the approach by weighting $tfn$ and removing the $ff$ component.

\begin{table}[htb]
\renewcommand\arraystretch{1.0}
\begin{center}
\small
\setlength{\tabcolsep}{0.4mm}{
\begin{tabular}{lrrrrrr}
\toprule
Method       & VI$\downarrow$        & mAP$\uparrow$                  & ARI$\uparrow$      & Dice$\uparrow$         & Betti$\downarrow$    \\ \midrule
w/o ff            & 39.52$_{\pm  4.38 }$          & 45.54$_{\pm  2.84 }$         & 63.59$_{\pm  2.32 }$      & 75.69$_{\pm 0.99}$        & 17.59$_{\pm 7.31}$      \\ 
w/o tt    & 39.67$_{\pm  3.76 }$            & 44.65$_{\pm  1.70 }$            & 63.80$_{\pm  0.74 }$         & 75.88$_{\pm 0.57}$    & 20.72$_{\pm 3.50}$      \\ 
w/o tt\&ff    & 48.04$_{\pm  2.55 }$            & 37.36$_{\pm  0.27 }$            & 59.07$_{\pm  0.58 }$     & 74.54$_{\pm 0.11}$       & 61.25$_{\pm 7.88}$            \\  
BoRT    & \textbf{36.59}$_{\pm  1.94 }$            &  \textbf{47.49}$_{\pm  0.88 }$           & \textbf{65.49}$_{\pm  1.21 }$        & \textbf{76.43}$_{\pm 0.26}$       & \textbf{12.72}$_{\pm 2.68}$      \\ \bottomrule
\end{tabular}}
\end{center}
\caption{Ablation study for loss components. The bold values indicate the best performance in each metric. BoRT means Skea-Topo here.}
\label{table2}
\end{table}

\subsubsection{The efficiency and effectiveness of BoRT}
To further demonstrate the efficiency and effectiveness of BoRT, thorough ablation experiments were compared to the advanced critical pixels detection-oriented Warp method. Specifically, we compared Warp, WaRT, and BoRT based on the Dice loss (identical to the Warp loss defined in \cite{warp}s) under identical conditions). WaRT employs the Warp approach to identify topological critical pixels and integrates the penalty term of BoRT into its loss function. Fig. \ref{fig6} (g) shows that BoRT delivers the best performance, and WaRT surpasses Warp in various metrics. Notably, the processing time of Warp is almost double that of BoRT. These results convincingly demonstrate that the method of BoRT to identify topological critical pixels is both efficient and effective. Furthermore, the penalty term introduced in BoRT proves more effective in promoting the network to preserve a topology. The comparison with Poly1, as shown in Fig. \ref{fig6} (f), further confirms the contribution of BoRT.

\subsubsection{Combining BoRT with Other Losses}
Our proposed BoRT serves as a lightweight plug-and-play loss function aimed at enhancing the effectiveness of various segmentation loss functions. To showcase its versatility, we integrated BoRT into five distinct loss functions: CE, Dice, BCE, ABW, and Skeaw. The results illustrated in Fig. \ref{fig6} demonstrate that incorporating BoRT leads to varying degrees of optimization, with improvements in the VI index ranging from 27.5 (CE) to 0.66 (Skeaw).

\section{Conclusion}

In this study, a Skea-Topo Aware loss was developed, tailored specifically to guide boundary segmentation networks towards the generation of topologically consistent results. It comprises a skeleton-aware weighted loss, which is enhanced to model geometric information more precisely by utilizing object skeletons, and a boundary rectification term, which imposes penalties on topologically critical pixels. These two components work together synergistically to improve segmentation performance. To efficiently identify the topologically critical pixels, a method based on the foreground and background skeletons was proposed. Our proposed method underwent rigorous evaluation through comparative and ablation experiments conducted on three datasets, demonstrating a maximum improvement of 7 points in the VI index, thereby validating its effectiveness and potential.

\noindent\textbf {Limitations} While extensively studied on 2D images, extending it to multi-class and 3D segmentation is theoretically feasible but requires further validation in future work.

\section*{Acknowledgments}
This work was supported by the National Key R $\&$ D Program of China under Grant 2022ZD0118001, the National Natural Science Foundation of China under Grant U22A2022 and 62106019, Scientific and Technological Innovation Foundation of Shunde Graduate School, USTB under Grant BK22BF010, and Fundamental Research Funds for the Central Universities of China under Grant 00007467, and China Postdoctoral Science Foundation under Grant 2021M700383. The computing work is supported by USTB MatCom of Beijing Advanced Innovation Center for Materials Genome Engineering.

\bibliographystyle{named}
\bibliography{ijcai24}

\end{document}